\documentclass{article}

\PassOptionsToPackage{square,numbers,sort}{natbib}

\usepackage[final]{neurips_2024}

\usepackage[utf8]{inputenc} 
\usepackage[T1]{fontenc}    
\usepackage[colorlinks=true, linkcolor=blue, citecolor=blue, urlcolor=blue]{hyperref}       
\usepackage{url}            
\usepackage{booktabs}       
\usepackage{amsfonts}       
\usepackage{nicefrac}       
\usepackage{microtype}      
\usepackage{xcolor}         
\usepackage[pdftex]{graphicx}       

\usepackage{amsmath}
\usepackage{cleveref}
\usepackage{fontawesome}
\usepackage{mathtools}
\usepackage[bb=boondox,bbscaled=.95,cal=boondoxo]{mathalfa} 

\usepackage{titlesec}
\titlespacing*{\section}{0pt}{18pt}{3pt}

\title{Future Research Avenues for Artificial Intelligence in Digital Gaming: An Exploratory Report}

\author{
	Markus Dablander\thanks{
		\href{https://scholar.google.com/citations?user=G-ufLUQAAAAJ\&hl=en}{\faGraduationCap\ Google Scholar} |
		\href{https://www.linkedin.com/in/markus-ferdinand-dablander}{\faLinkedinSquare\ LinkedIn} | \href{https://github.com/MarkusFerdinandDablander}{\faGithub\ GitHub} }\vspace{0pt} \\
	Commissioned by Beam Foundation\thanks{\href{https://onbeam.com}{\faHome\ Homepage} | \href{https://x.com/BuildOnBeam}{$\mathbb{X}$}} \ as Independent Contract Artificial Intelligence Researcher  \\
}

\begin{document}

\maketitle

\vspace{-10pt}

\begin{figure}[h]
	\centering
	\includegraphics[width=0.5\linewidth]{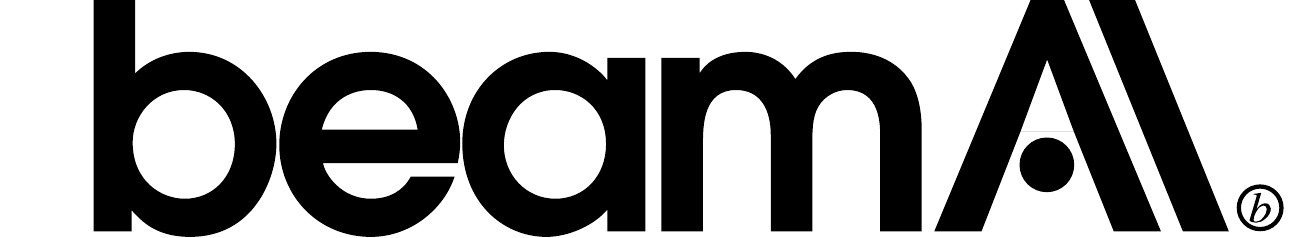}
	\label{fig:beamlogo}
\end{figure}

\vspace{70pt}

\begin{abstract}

Video games are a natural and synergistic application domain for artificial intelligence~(AI) systems, offering both the potential to enhance player experience and immersion, as well as providing valuable benchmarks and virtual environments to advance AI technologies in general. This report presents a high-level overview of five promising research pathways for applying state-of-the-art AI methods, particularly deep learning, to digital gaming within the context of the current research landscape. The objective of this work is to outline a curated, non-exhaustive list of encouraging research directions at the intersection of AI and video games that may serve to inspire more rigorous and comprehensive research efforts in the future. We discuss (i)~investigating large language models as core engines for game agent modelling, (ii)~using neural cellular automata for procedural game content generation, (iii)~accelerating computationally expensive in-game simulations via deep surrogate modelling, (iv)~leveraging self-supervised learning to obtain useful video game state embeddings, and (v)~training generative models of interactive worlds using unlabelled video data. We also briefly address current technical challenges associated with the integration of advanced deep learning systems into video game development, and indicate key areas where further progress is likely to be beneficial.

\end{abstract}

\newpage

\renewcommand\thefootnote{\arabic{footnote}}
\setcounter{footnote}{0}
\setcounter{page}{2}

\section{Introduction}

In the last decade, the rise of advanced neural network architectures has led to a series of dramatic breakthroughs in the fields of machine learning and artificial intelligence~(AI). The GPU-accelerated training of large, carefully designed deep learning models has enabled researchers to tackle previously intractable challenges in diverse areas such as computer vision~\citep{krizhevsky2012imagenet, simonyan2014very, zeiler2014visualizing, szegedy2015going, he2016deep}, natural language processing~\citep{wu2016google, vaswani2017attention, brown2020language, raffel2020exploring}, artificial content generation~\citep{kingma2013auto, goodfellow2014generative, rombach2022high, ho2020denoising}, and computational chemistry~\citep{gilmer2017neural, gomez2018automatic, duvenaud2015convolutional, xu2018powerful, jumper2021highly}. One exceptionally promising and natural application area for modern deep learning, which will be explored in this report, is \textbf{digital gaming}.

The focus of AI research on games already has a long and important history. In particular, the study of classical board games such as Chess, Checkers, and Go has been formative and instrumental for the AI field as a whole~\citep{hu2024games, yannakakis2018artificial}. The highly structured nature of many games allows for the emergence of great complexity and strategic depth from simple rules that can easily be expressed in a computational framework; consequently, games have long been considered ideal testing grounds for the reasoning and planning capabilities of AI agents. A significant milestone was reached in $2016$, when the first AI system achieved superhuman performance in the game of Go~\citep{silver2016mastering}, which, at that time, represented the last major, popular board game in which human experts still outperformed computers.

Particularly since then, digital games have increasingly been recognised as one of the next great frontiers of AI research. In recent years, considerable progress has been made towards developing AI agents capable of mastering real-time strategy video games, such as StarCraft~II~\citep{vinyals2019grandmaster}, and multiplayer online battle arena video games, such as Dota 2~\citep{berner2019dota}, both of which pose a far greater challenge to AI systems than classical board games. Simultaneously, the construction of \textit{general} AI models that can learn to play multiple, qualitatively distinct arcade video games has emerged as an active field of research~\citep{lee2022multi, reed2022generalist, mnih2015human}, and work in this area may serve as a stepping stone towards the development of more general AI systems in other domains.

Importantly, it is not merely the case that video games have the potential to enrich contemporary AI research; the converse is true as well. The relationship between AI research and digital gaming is mutual and synergistic~\citep{yannakakis2018artificial, miikkulainen2006computational, hu2024games}, with video games providing valuable benchmarks, test-beds and virtual environments for novel AI systems, while novel AI systems, in turn, provide a wealth of opportunities for video game developers to enhance their creative products. Partly due to the rapid progress of recent AI technologies, in particular deep learning, many of these opportunities are still underutilised and have yet to be explored.

This report aims to give a concise, preliminary overview of a selection of five potential research avenues for the application of state-of-the-art AI techniques to digital gaming. While our emphasis will mainly be on research directions where contemporary AI methods can enhance digital gaming, the reciprocal connection between AI and video games makes it conceivable that investigating these topics could also drive new insights and advancements in AI itself. The objective of this exploratory report is not to provide a comprehensive set of mature research proposals, or to present novel original research findings. Instead, the focus is on offering a speculative collection of high-level ideas that may serve to inspire more rigorous and focused research efforts in the future. The selected areas are not in any way exhaustive, but rather represent a curated and necessarily subjective collection of ideas deemed particularly intriguing during our examination of the current research landscape.

The foundational book from~\citet{yannakakis2018artificial}, which served as one of the most valuable references for this work, outlines three core applications of AI to video gaming:

\begin{itemize}
	\item \textbf{AI for game playing and agent modelling}, which includes simulating the role of a human player~\citep{vinyals2019grandmaster, berner2019dota}, or controlling other game agents in the broadest sense, such as non-player characters~(NPC)~\citep{orkin2006three, aversa2013action}, or hidden agents governing aspects of the game environment~\citep{booth2009ai}.
	
	\item \textbf{AI for procedural content generation}~\citep{shaker2016procedural}, which includes the algorithmic creation of game levels, music, textures, art, dialogues, items, characters, or any other digital content.
	
	\item \textbf{AI for player modelling}~\citep{yannakakis2013player}, which includes the modelling of human player characteristics, such as player type, predicted in-game behaviour, or emotional state, based on measured gameplay and player data.
\end{itemize}

All of these three areas are reflected in the research avenues discussed below, with a greater emphasis on the first two.

\section{Large Language Models for Game Agent Modelling}

Large language models~(LLMs) such as OpenAI's GPT-4~\citep{achiam2023gpt}, Google's Llama~3~\citep{dubey2024llama}, and Anthropic's Claude~3~\citep{enis2024llm} have recently risen to enormous prominence due to their advanced capabilities to maintain realistic conversational arcs and generate flexible solutions to a wide range of language-related tasks. Current state-of-the-art LLMs are based almost exclusively on variants of the transformer architecture, introduced in the seminal work of~\citet{vaswani2017attention} in $2017$. Transformer networks rely on the concept of \textit{self-attention}, a deep learning mechanism designed to effectively capture long-range dependencies and contextual information in sequential data. 
The core training process for many LLMs is self-supervised and autoregressive, meaning that the LLM is trained to generate text by probabilistically predicting the next word (or subword token) in a text based on the preceding words. LLMs regularly contain billions of trainable parameters and are frequently trained on vast corpora of unlabelled textual data collected from publicly available sources such as books and websites~\citep{du2024survey}.

At the moment, LLMs are attracting significant attention within the video game AI research community for their potential applicability to a diverse array of gaming-related tasks~\citep{gallotta2024large, sweetser2024large}. For example, LLMs have recently been explored for the algorithmic creation of new video game levels in \textit{Super Mario Bros}~\citep{sudhakaran2024mariogpt}, the autonomous playing of \textit{Minecraft} through the generation of code for a suitable game API~\citep{wang2023voyager}, the systematic extraction of player sentiment from written game reviews~\citep{viggiato2023leveraging}, and the automatic generation of dynamic audio commentary for \textit{League of Legends} gameplay~\citep{ranella2023towards}. Covering all promising use cases of LLMs in digital gaming would be beyond the scope of this exploratory report. However, we briefly highlight one possible research direction we consider to be particularly interesting, namely the use of LLMs for \textbf{game agent modelling}.

Game agent modelling includes the development and control of NPCs such as teammates, enemies, sidekick companions, merchants, bystanders, and other virtual characters in the broadest sense. Perhaps one of the most evident and fruitful applications of LLMs in this context would be to equip NPC agents with the ability to have natural and unscripted conversations with each other and with human players. First investigations in this area have already begun~\citep{muller2023chatter, park2023generative, nananukul2024if}; further advancements in integrating LLMs as NPC dialogue systems may be able to markedly enhance the realism of virtual characters, leading to substantially deeper and more immersive video game experiences.

However, the overall potential of LLMs for agent modelling may exceed the already appealing area of dynamic dialogue generation. In $2024$, \citet{hu2024survey} gave a conceptual description of an entire cognitive architecture for general game agents that embeds an LLM as the core thinking component within a network of other submodules covering perception, memory, role-playing, action, and learning. Drawing closely from the work of Hu et al.,~one might envision an LLM-based cognitive architecture broadly working as follows: the \textbf{perception} module translates current game states into textual descriptions; the \textbf{thinking} module, powered by an LLM, receives outputs from the perception module and relevant text-based memories retrieved from the \textbf{memory} module to output textual action plans; these plans are translated by the \textbf{action} module into executable low-level in-game actions; the LLM-based thinking process is additionally biased with character information by the \textbf{role-playing} module; and continuously updated with techniques such as reinforcement learning or supervised finetuning by the \textbf{learning} module. One may also consider introducing a separate \textbf{goal} module that manages the objectives of the agent in a text-based manner and interacts with the other modules.

\begin{figure}
	\centering
	\includegraphics[width=0.95\linewidth]{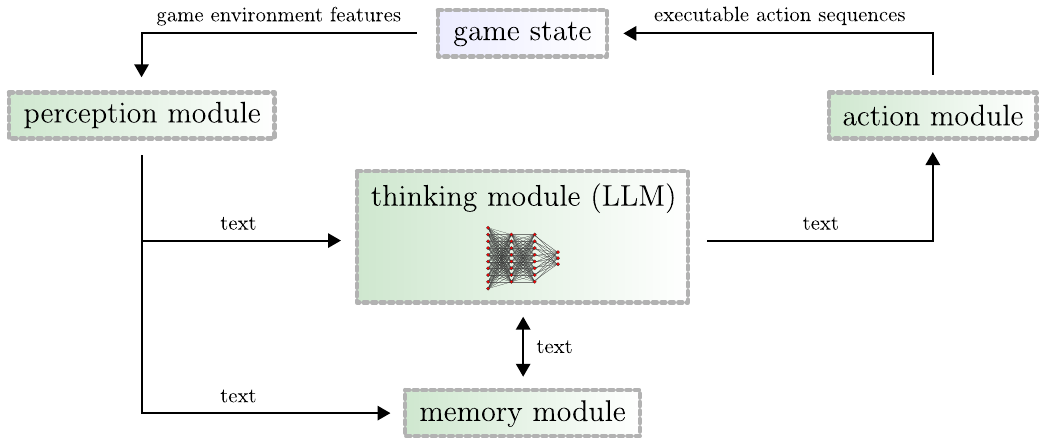}
	\caption{Simple, high-level overview of a conceivable LLM-based cognitive architecture for a video game agent, strongly influenced by the works of~\citet{park2023generative} and~\citet{hu2024survey}. The perception module translates game environment features (pixels, statistical features, vectorial embeddings, etc.) extracted from the game state into textual descriptions. The memory module stores past textual perceptions, as well as other memory items that are either predetermined (fixed character information, basic goals, etc.) or generated by the thinking module (novel knowledge, reflections, goals, procedural skills, etc.). The thinking module, based on a large language model~(LLM), processes current textual perceptions and relevant textual memory items retrieved from the memory module, and outputs textual action plans and new memory items. The textual action plans are converted by the action module into low-level sequences of in-game behaviours that are executed to change the game state.}
	\label{fig:llmagents}
\end{figure}

While each of the above modules could easily warrant its own extensive research programme, first successful attempts to design game agents via the integration of LLMs into broader cognitive architectures have already been made. Most notably,~\citet{park2023generative} created an interactive artificial society consisting of a virtual 2D village with $25$ distinct LLM-based game agents with different personalities and professions. Each agent maintains a text-based memory stream that contains a comprehensive list of the agent's perceptions, along with generated action plans and synthesised higher-order reflections. An LLM interacts with the agent's memory stream and current perceptions to generate new reflections and adapt action plans. This approach leads to an impressively complex and convincing set of self-organising emergent social behaviours: agents lead natural dialogues, coordinate actions, spread information, and dynamically update social relationship memories. A simple, schematic overview of an LLM-based cognitive architecture, heavily inspired by the works of~\citet{park2023generative} and~\citet{hu2024survey}, is depicted in~\Cref{fig:llmagents}.

Further efforts, such as those by Park et al., to integrate LLMs into a broader network of cognitive modules could not only contribute to the development of more immersive video game characters and more human-like virtual agents for playtesting, but also advance research on the disputed question of how suitable LLMs truly are as core engines for artificial general intelligence~\citep{goertzel2023generative, bubeck2023sparks}.

\section{Neural Cellular Automata for Procedural Content Generation}

Cellular automata~(CA)~\citep{von1966theory, sarkar2000brief} are a family of extensively investigated and diverse mathematical models represented by grids of cells, whose states evolve in discrete time. At each time point $t$, each cell has a state represented by a number (or a vector of numbers), and its state at time $t+1$ is determined by its own state and the states of its neighboring cells at time $t$, according to a local transition function that defines how the states evolve.

A simple and iconic example of CA that many readers may be familiar with is given by \textit{Conway's Game of Life}~\citep{gardner1970mathematical}, which takes place on an infinite 2D orthogonal grid of square cells, each of which can only be in one of two possible states, \textit{dead} or \textit{alive}. Given some initial configuration, cell states start to evolve based on a simple transition function that only takes into account how many dead or alive neighbours a cell has at a given time. In spite of its extreme simplicity, Conway's Game of Life exhibits an impressive set of complex self-organising behaviours.\footnote{\href{https://www.youtube.com/watch?v=C2vgICfQawE}{Video illustration of Conway's Game of Life}}

CA have already been used in video games with considerable success, for instance to grow infinite cave levels for the game \textit{Cave Crawler}~\citep{johnson2010cellular}, automatically generate playable mazes for maze running games~\citep{adams2017procedural}, model granular media like sand or soil~\citep{devlin2021probabilistic}, or simulate erosion in virtual environments~\citep{olsen2004realtime}. CA are highly computationally efficient models that can be used to generate intricate virtual content. At the same time, CA are conceptually simple, intuitive to understand and easy to implement. However, the constructive, emergent nature of CA also makes them difficult to control~\citep{yannakakis2018artificial}. In general, given a local transition function, it is very difficult to predict which pattern will arise over time from a specific initial grid state; even extremely similar initial states may quickly diverge in a chaotic manner, leading to entirely different outcomes~\citep{urias1997sensitive}. Similarly, identifying a local transition function that over time maps a given initial state to a desired pattern is a nontrivial technical problem. These properties limit the utility of CA as procedural content generators for video games by making it challenging to impose essential constraints on generated content. Such constraints may include guaranteed solvability for a game level, or a particular shape, connectivity and aesthetic style for a game object.

\begin{figure}
	\centering
	\includegraphics[width=0.95\linewidth]{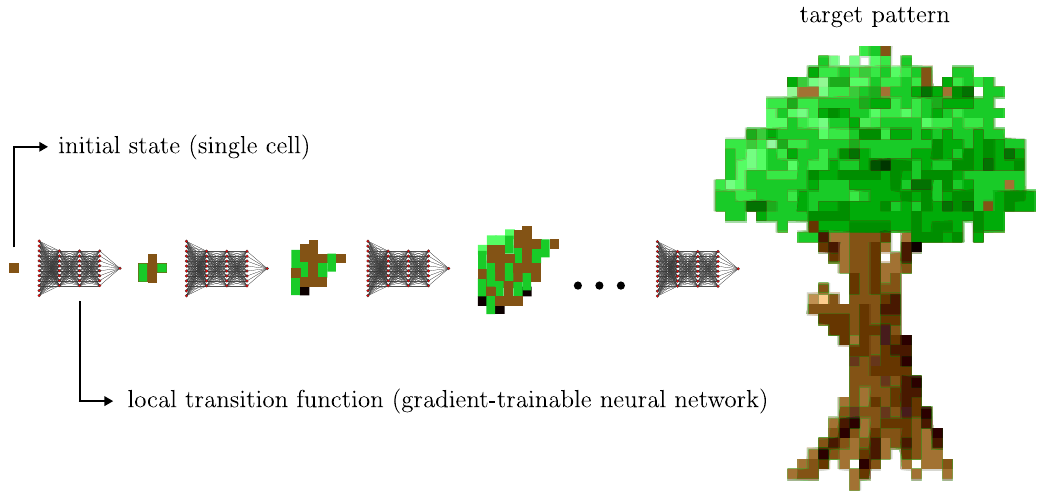}
	\caption{Conceptual diagram of how a neural cellular automaton (NCA), once trained, could iteratively generate the target image of a tree (image not generated by actual NCA, used for illustrative purposes only). An NCA is a cellular automaton whose local transition function is parametrised by a neural network. \citet{mordvintsev2020growing} showed how an NCA can be trained with gradient-based methods to organically grow an arbitrary, predefined target pattern from a single initial cell. The NCA can also learn to automatically converge back to its intended target pattern when disturbed in a manner that resembles \textit{self-regeneration}.}
	\label{fig:ncas}
\end{figure}

Recently, \textbf{neural cellular automata~(NCA)}~\citep{wulff1992learning, nichele2017neat} have been increasingly investigated as a significant way to address some of these shortcomings and allow for substantially greater control over the dynamical processes governing CA. An NCA is a CA whose local transition function is parametrised by a trainable neural network. One of the key contributions in the field of NCA was made by~\citet{mordvintsev2020growing} in $2020$, who demonstrated how an NCA parametrised by a convolutional neural network can be effectively trained in a differentiable end-to-end manner via gradient-based methods to iteratively generate any predefined target image from a single cell (see~\Cref{fig:ncas} for an illustration of this idea).
They moreover showed how NCA can be trained to exhibit \textit{self-regeneration}, or, in the language of dynamical systems theory, how the target image can be turned into an attractor. A self-regenerating NCA automatically converges back to its intended target pattern when perturbed.\footnote{\href{https://distill.pub/2020/growing-ca}{Interactive animations of self-regenerating NCA by~\citet{mordvintsev2020growing}}} 

The seminal work of~\citet{mordvintsev2020growing} has implications stretching into diverse areas, including morphogenesis, embryonic development, regenerative medicine, self-organisation, and swarm robotics. In addition, first attempts have already been made to apply NCA in video game research~\citep{earle2022illuminating, mordvintsev2021mu, pajouheshgar2023dynca, sudhakaran2021growing, pajouheshgar2024mesh, sato2023automata}: \citet{earle2022illuminating} successfully trained NCA to generate levels for 2D tile-based games while taking into account validity and diversity constraints; \citet{sudhakaran2021growing} used NCA in the virtual world of \textit{Minecraft} for the targeted morphogenetic growth of complex 3D objects such as castles and trees; and~\citet{pajouheshgar2024mesh} employed NCA for the virtual synthesis of desired textures on 3D meshes.

These early studies highlight the potential of NCA as a novel deep-learning-based tool for procedural content generation in virtual environments~\citep{liu2021deep}. However, many promising research directions remain to be investigated. Additional work could further explore the capabilities of NCA to be trained via custom loss functions designed to promote specific design constraints for video game content generation. Future studies could also more deeply investigate NCA for the creation of realistic textures for digital objects in a computationally efficient manner, or for accurately simulating organic and regenerative processes, such as the growth of natural ecosystems, aging characters, material degradation, or wound healing. Beyond content generation, it may also be interesting to investigate NCA as efficiently trainable swarm intelligence models to induce emergent behaviours in groups of locally connected NPCs.

\section{Deep Surrogate Modelling to Accelerate Computationally Expensive In-Game Simulations}

In their seminal study,~\citet{gilmer2017neural} not only introduced \textit{message-passing} as a unifying framework for graph neural network architectures; in addition to this salient contribution, they also showed that a graph neural network can learn to efficiently predict quantum-chemical properties of small molecules using the QM9 data set~\citep{ramakrishnan2014quantum} for training. The QM9 data set consists of around $134$k chemical compounds; each compound comes with a set of numerical labels that represent approximations of relevant quantum-chemical properties. For each compound, each numerical label is the result of a quantum-mechanical simulation based on what is known as \textit{density functional theory}~\citep{becke1992density}. Density functional theory simulations, while highly useful for elucidating the electronic structure of a molecule, are associated with prohibitive computational costs; for example, generating all the labels in the QM9 data set for a single molecule with nine heavy atoms on a single core of a Xeon E5-2660 processor with 2.2 GHz and commonly used software may take around one hour~\citep{gilmer2017neural}. In contrast, the graph neural network from Gilmer~et~al., which was trained on the QM9 data set in a supervised manner, can estimate the outcome of density functional theory simulations for a novel molecule in a fraction of a second. This corresponds to a speed-up of five orders of magnitude, making it computationally feasible to rapidly predict quantum-chemical properties for large molecular libraries.

\begin{figure}[h]
	\centering
	\includegraphics[width=0.85\linewidth]{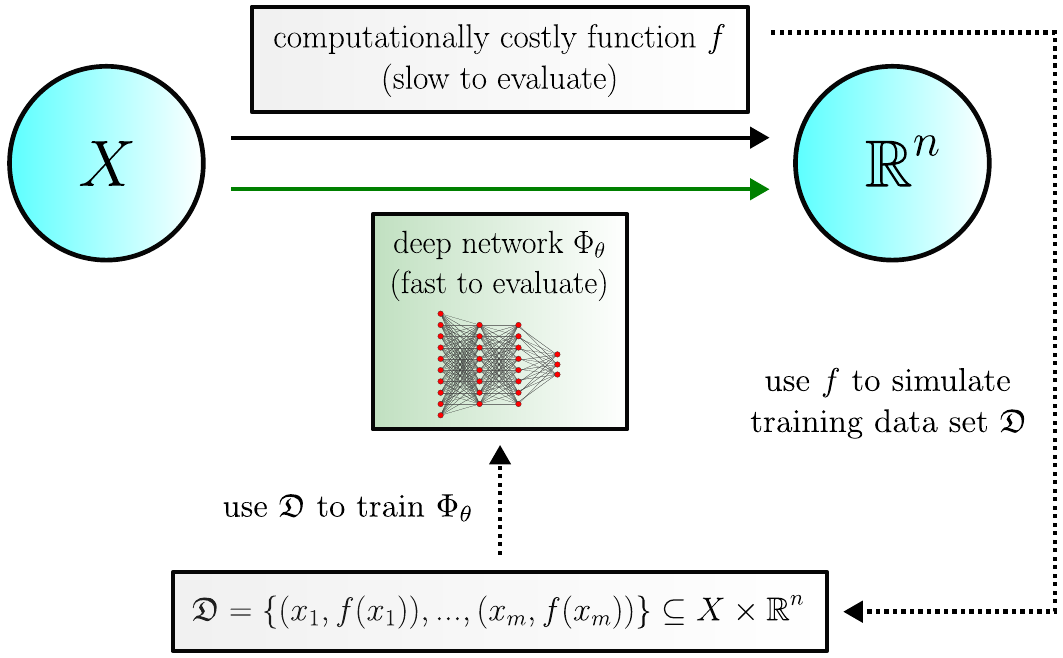}
	\caption{Illustration of the elementary idea behind deep surrogate modelling: A computationally expensive function $f$ is repeatedly evaluated to generate a training data set $\mathfrak{D}$, which is then used to train a deep network~$\Phi_{\theta}$. After training, $\Phi_{\theta}$ acts as a computationally fast approximation of $f$.}
	\label{fig:deppsurrmod}
\end{figure}

The work of Gilmer~et~al.~represents a prime example of what can be referred to as \textbf{deep surrogate modelling}~\citep{sudret2017surrogate, zhu2019physics, tang2020deep, radaideh2020surrogate}. A high-level illustration of the key idea behind deep surrogate modelling is given in~\Cref{fig:deppsurrmod}. In one of its most elementary forms, deep surrogate modelling is a technique used to speed up the evaluation of a computationally expensive function
$$f : X \to \mathbb{R}^n$$
that is of interest for a practical application. For instance, $f$ could represent an expensive numerical computer simulation. In our earlier example from Gilmer~et~al., the domain $X$ would be a set of molecular graphs, and $f$ would represent a density functional theory simulation that maps molecular graphs to numerical quantum-chemical properties expressed as vectors in $\mathbb{R}^n$. Initially, a (sometimes considerable) computational effort is made to create a data set
$$\mathfrak{D} = \{(x_1, f(x_1)), ..., (x_m, f(x_m)) \} \subseteq  X \times \mathbb{R}^n,$$
which is then used to train the parameters $\theta$ of a suitable deep learning architecture
$$\Phi_{\theta} : X \to \mathbb{R}^n$$
in a supervised manner. After training, the deep network $\Phi_{\theta}$ can be used as a \textit{surrogate} for $f$, approximating the value of $f(x)$ with $\Phi_{\theta}(x)$ for novel $x$ outside the training set $\mathfrak{D}$. Furthermore, $\Phi_{\theta}$ can also be optimised instead of $f$ when looking for maximisers or minimisers of $f$. While $\Phi_{\theta}$ may be less accurate than the original simulation function $f$, it can be orders of magnitude faster to evaluate.

Deep surrogate modelling is particularly useful in situations requiring computationally expensive and repetitive simulations~\citep{zhu2019physics}. Video games regularly involve a plethora of such simulations, spanning diverse areas like gameplay balancing, difficulty tuning, fluid and particle dynamics, Newtonian mechanics, pathfinding, environmental systems, realistic lighting, sound propagation, game state prediction, and procedural content generation. As such, digital gaming may be well-suited for the application of deep surrogate models to accelerate gameplay, reduce loading times and optimise the game development process. Despite these encouraging possibilities, the number of studies exploring deep surrogate models for video games appears to be relatively limited~\citep{karavolos2019multifaceted, bhatt2022deep, karavolos2018pairing, karavolos2018using, zhang2022deep, migkotzidis2021susketch, karavolos2017learning}. However, early work in this field has already shown some success. For example, \citet{bhatt2022deep} trained a deep surrogate model on simulated data to predict the behaviour of a game agent in novel environments, applying the model to accelerate the algorithmic generation of new environments that lead to diverse agent behaviours. Overall, the utility of deep surrogate modelling for digital gaming may still be underexplored, offering notable opportunities for future research.

\section{Self-Supervised Video Game State Representation Learning}

Being able to represent the abstract \textbf{state} of a video game in terms of a meaningful numerical vector is a key element in a large variety of modern AI applications for digital gaming~\citep{yannakakis2018artificial}. In this context, a high-quality vectorial representation technique should be able to condense the essential features of a video game state into an informative embedding that can be effectively used for downstream AI tasks. Such tasks may, for example, include using a game state embedding as a model of perception for an autonomous game agent~\citep{anand2019unsupervised, mnih2015human}, predicting the emotional state of a player from gameplay video streams~\citep{makantasis2019pixels}, predicting future game states from current ones~\citep{oh2015action}, dynamically adapting game music depending on the state of a game~\citep{plut2020generative}, or algorithmically translating game states into natural language descriptions~\citep{park2023generative}. 

A powerful deep learning paradigm for vectorial data representation that has emerged in recent years is \textbf{self-supervised learning}~\citep{liu2021self}, which offers a collection of strategies to learn rich, general-purpose embeddings solely from the internal structure of unlabelled input data. While supervised deep learning is based on the extraction of task-specific features from labelled data sets, self-supervised learning does not rely on data annotation by human subjects, and instead allows one to find flexible feature representations in a task-agnostic manner. Labelled data is frequently scarce and hard to obtain; self-supervised learning methods do not suffer from comparable limits, as they can take advantage of vast corpora of unlabelled data sets, such as curated libraries of images and text extracted from the internet. Representations learned via self-supervised training can be employed in a variety of ways, including clustering~\citep{sharma2019self} and anomaly detection~\citep{yoa2021self}. Importantly, they can also be fine-tuned on downstream supervised tasks~\citep{chen2020big}, an approach that regularly leads to substantial boosts in performance compared to purely supervised techniques.

While self-supervised learning has become a significant area of research in domains like natural language processing and computer vision~\citep{jaiswal2020survey}, comparable work in digital gaming is still relatively sparse. A literature search for studies that use concepts from self-supervised learning in digital gaming revealed only $12$ instances~\citep{trivedi2022learning, anand2019unsupervised, trivedi2023towards, trivedi2021contrastive, lee2022self, terbouche2022comparing, ravsajski2024behave, trivedi2022game, team2021creating, basavatia2024starling, khameneh2020entity, laskin2020curl}. In a notable article,~\citet{anand2019unsupervised} introduced a systematic benchmark to evaluate self-supervised learning methods for video game states via the prediction of essential internal game variables in Atari $2600$ games from learned representations. They employed this benchmark to demonstrate the effectiveness of a mutual-information-based representation learning strategy. In a related study,~\citet{trivedi2022learning} showed that three popular self-supervised learning strategies applied to video game pixels alone can be used to derive game state embeddings that are predictive of key internal game variables, such as enemy positions on the screen in a first-person shooter, or game world coordinates of football players and the ball in a football simulator.

\begin{figure}
	\centering
	\includegraphics[width=0.95\linewidth]{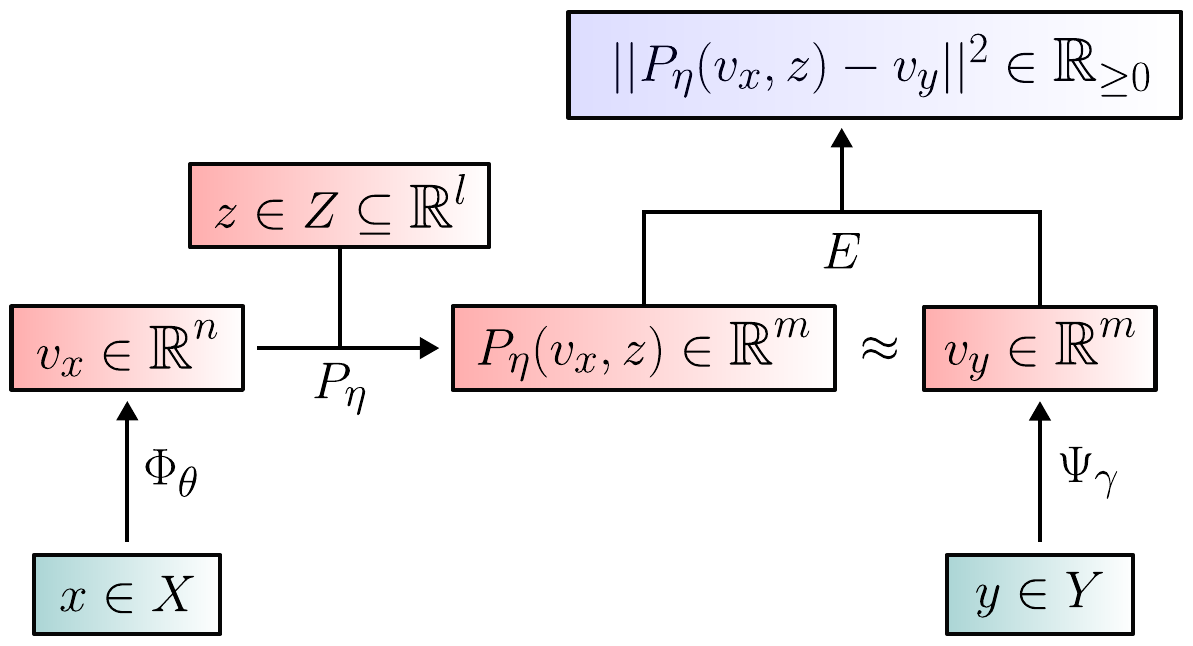}
	\caption{Schematic overview of a prototypical joint-embedding predictive architecture~(JEPA)~\citep{lecun2022path} for self-supervised learning. The variables $x$ and $y$ could, for instance, represent images of game pixels at times $t$ and $t + \delta$, the encoders $\Phi_{\theta}$ and $\Psi_{\gamma}$ could be convolutional neural networks that map the images to embeddings $v_x, v_y$, the latent variable $z$ could symbolise the action taken by the player at time~$t$, and $P_{\eta}$ could be a multilayer perceptron whose output $P_{\eta}(v_x, z)$ aims to approximate $v_y$.}
	\label{fig:jepa}
\end{figure}

Representing video game states via state-of-the-art self-supervised learning may be an impactful area for future research. One particularly interesting approach could be to further investigate \textit{joint-embedding predictive architectures}~(JEPAs)~\citep{lecun2022path, dawid2024introduction} for this purpose. JEPAs constitute a novel self-supervised learning framework with attractive technical properties that has recently demonstrated encouraging results in the image domain~\citep{assran2023self}. Let 
$$\Psi_{\gamma} : Y \to \mathbb{R}^m, \quad \Phi_{\theta}: X \to \mathbb{R}^n,$$ 
be two trainable deep learning encoders that map given data entities in the sets $X$ and $Y$ (which could, for example, be collections of images or graphs) to vectors. In essence, a JEPA aims to learn useful representations by training to predict the embedding 
$$v_y \coloneqq \Psi_{\gamma}(y) \in \mathbb{R}^m$$ 
of an input entity $y \in Y$ from the embedding 
$$v_x \coloneqq \Phi_{\theta}(x) \in \mathbb{R}^n$$ 
of a somehow related input entity $x \in X$, with the help of a latent variable
$$z \in Z \subseteq \mathbb{R}^l  $$ that can be used to add additional information about $y$ not contained in $x$. The prediction of $v_y$ from $v_x$ and $z$ is done via a trainable predictor 
$$P_{\eta}: \mathbb{R}^n \times Z \to \mathbb{R}^m $$ 
whose aim is to minimise a scalar error function such as
$$E(P_{\eta}(v_x, z), v_y) \coloneqq \vert \vert P_{\eta}(v_x, z) - v_y \vert \vert^2 \in \mathbb{R}_{\geq 0}.$$
A schematic visualisation of this architecture is depicted in~\Cref{fig:jepa}.

Since the prediction of $y$ from $x$ and $z$ occurs implicitly in abstract representation space, unimportant details of $x$ and $y$ can be eliminated by the encoders $\Psi_{\gamma}$ and $\Phi_{\theta}$ prior to prediction. This architecture also allows for multiple values of $y$ to be compatible with a single value of $x$, due to potential invariance properties of the encoder $\Psi_{\gamma}$ and the ability to change the output of $P_{\eta}$ by varying the latent variable $z$.

In the context of video games, $x$ and $y$ could, for example, represent game pixels on the player screen at times $t$ and $t + \delta$, the encoders $\Phi_{\theta}$ and $\Psi_{\gamma}$ could be convolutional neural networks that map the pixels to vectorial embeddings $v_x, v_y$, the latent variable $z$ could symbolise the action taken by the player at time~$t$, and $P_{\eta}$ could be a multilayer perceptron whose output $P_{\eta}(v_x, z)$ aims to approximate $v_y$. In other words, $P_{\eta}$ could be trained to predict the abstract future game state from the abstract current game state given the player action, a task that may encourage $\Phi_{\theta}$ and $\Psi_{\gamma}$ to simultaneously learn meaningful abstract game state embeddings. The BYOL method~\citep{grill2020bootstrap}, which shares some similarities with the JEPA approach, has already been successfully tested for pixel-based game state representation learning in the previously mentioned study by~\citet{trivedi2022learning}. For a more detailed technical description of JEPAs, including important training considerations to prevent such architectures from collapsing into producing only constant representations, we refer the reader to the original article by~\citet{lecun2022path}.

\section{Learning Generative Models of Interactive Worlds from Unlabelled Videos}

In early $2024$, Google DeepMind introduced \textbf{Genie}~\citep{bruce2024genie}, an $11$-billion parameter generative world model trained in a self-supervised manner on a large-scale library of publicly available gameplay videos of 2D platformer games. Genie can automatically generate an infinite variety of novel and action-controllable 2D platformer gaming worlds.\footnote{\href{https://sites.google.com/view/genie-2024/}{Video illustrations of Genie's~\citep{bruce2024genie} capabilities}} Each unique world is created using only a single image as an initial seed.

The neural architecture of Genie consists of three major components, all of which rely on the use of computationally efficient spatiotemporal transformer models~\citep{xu2020spatial, vaswani2017attention}: a video tokeniser, a latent action model, and a dynamics model. The video tokeniser is implemented via a VQ-VAE~\citep{van2017neural} that is trained to translate video game frames into discrete vectorial tokens, and vice versa. The latent action model is trained to infer plausible actions between consecutive pairs of frames. It too is based on a VQ-VAE architecture that naturally allows for limiting the number of possible actions to a small, fixed-size set of discrete vectorial action embeddings. Most notably, the latent action model is trained in an entirely self-supervised way, without the need for human-annotated action labels. The dynamics model~\citep{chang2022maskgit} is trained autoregressively to predict future tokenised frames from past and current tokenised frames and inferred action embeddings. As a result, the dynamics model is encouraged to learn the consequences of actions on the temporal evolution of the video game world.

\begin{figure}
	\centering
	\includegraphics[width=1\linewidth]{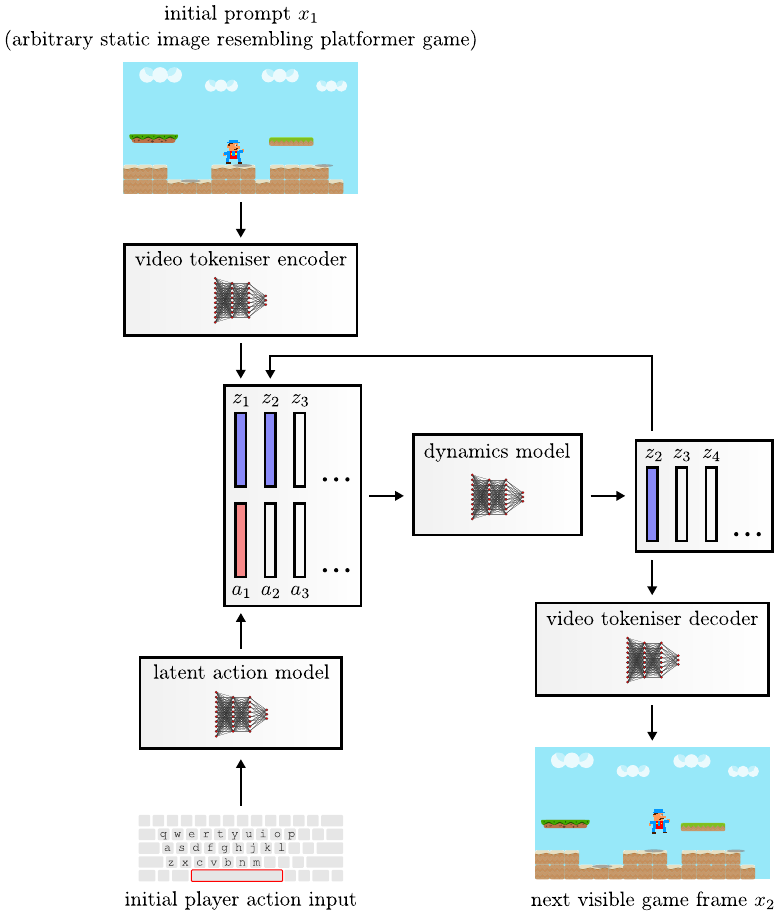}
	\caption{Schematic diagram illustrating the inference process of the trained Genie model~\citep{bruce2024genie} to generate a playable platformer game from a given image prompt~$x_1$ (images not generated by actual Genie system, used for illustrative purposes only). The video tokeniser and the latent action model respectively translate the prompt image~$x_1$ and the initial player action input into embeddings~$z_1$ and~$a_1$, which are subsequently used by the dynamics model to predict the next tokenised frame~$z_2$. The compressed representation $z_2$ is then converted by the video tokeniser into a visible game frame~$x_2$. This process is iteratively repeated using previously generated image tokens and recorded input actions to give rise to a sequence of interactive game frames~$(x_1, x_2, x_3, ...)$. }
	\label{fig:genie}
\end{figure}

Once trained, the Genie system can be operated in the following way: 
\begin{enumerate}

\item A user first prompts Genie with an image $x_1$ vaguely resembling a scene from a platformer game. This image serves as the initial game frame. The starting image $x_1$ could for instance be a screenshot from an actual game, an imagined sketch drawn by a human, or an artificial image created via a text-to-image generator~\citep{saharia2022photorealistic} from a natural language description.

\item The image $x_1$ is compressed into a discrete vectorial token $z_1$ using the video tokeniser. 

\item The player can then input an initial action which is translated into a discrete vectorial action embedding~$a_1$ by a component of the latent action model.

\item The dynamics model uses its acquired world knowledge~\citep{ha2018world} to predict the next tokenised frame $z_2$ based on action $a_1$ and state $z_1$. 

\item The compressed vectorial token $z_2$ is decoded into the next video game frame $x_2$ by the video tokeniser. The image frame $x_2$ is displayed to the user.

\item The last three steps are iteratively repeated to give rise to an interactive sequence of image frames
$$(x_1, x_2, x_3, ...) $$ 
that constitute a playable platformer game. For example, after the initial iteration, the user specifies another action $a_2$, the dynamics model uses $(z_1, z_2)$ and $(a_1, a_2)$ to predict $z_3$, and $z_3$ is subsequently decoded into the next visible frame $x_3$.

\end{enumerate}
This process is visualised in~\Cref{fig:genie}.

One of the limitations of Genie outlined in the original article~\citep{bruce2024genie} is its low frame rate, reported to be around one frame per second. Genie may also sometimes hallucinate unrealistic future scenarios, or fail to maintain the stability and consistency of a generated world over time. Despite these shortcomings, Genie's architecture represents a strong proof of concept for the possibility of learning \textbf{generative models of interactive worlds} from gameplay video data alone. The fact that Genie is able to effectively infer an operable latent action space entirely without human-annotated action labels is particularly noteworthy.

Note that this article is written only a few days after Google DeepMind's public announcement of Genie~2~\citep{deepmind2024genie2}. Due to the early stage of this research, a comprehensive technical description of this novel model via an associated research paper is currently lacking. It has been stated, however, that Genie~2 is an autoregressive latent diffusion model~\citep{rombach2022high}. It appears that Genie~2 extends the domain of Genie to the substantially more challenging task of generating complex, interactive 3D game worlds\footnote{\href{https://deepmind.google/discover/blog/genie-2-a-large-scale-foundation-world-model}{Video illustrations of Genie~2's capabilities}} instead of simple 2D platformer games. Genie~2 is reported to exhibit a set of emergent capabilities related to physics, object interactions, character animation and game agents, and to be able to maintain a consistent 3D world for up to a minute.

The potential implications of interactive-world generators like Genie and Genie 2 for general AI research as well as digital gaming are noteworthy. In the future, considerably more mature versions of such systems could conceivably become useful for a wide array of tasks, including training and testing general adaptive agents in maximally diverse virtual environments~\citep{raad2024scaling}, simulating an infinite stream of realistic training scenarios for robotic systems like autonomous vehicles~\citep{kaur2021survey}, accelerating video game development via rapid prototyping, allowing non-experts to easily create their own action-controllable video game snippets, procedurally extending existing video games and virtual worlds~\citep{liu2021deep, summerville2016super}, automatically personalising video game content based on a model of player behaviour or characteristics~\citep{yannakakis2013player}, and simulating virtual environments for human training purposes in fields like medicine~\citep{ruthenbeck2015virtual} and aviation~\citep{lee2017flight}. 

Future impactful work in this area could for example focus on maintaining the consistency and stability of generated environments over extended periods of time, the prevention of hallucinations~\citep{andriopoulos2023augmenting, rawte2023troubling}, the inclusion of audio signals, the generation of game mechanics for less explored genres like bird's-eye strategy games, the identification and mitigation of computational bottlenecks~\citep{menghani2023efficient} to accelerate low frame rates, and the development of even more capable dynamics and latent action models~\citep{ye2024latent}. In addition, it may be interesting to investigate to what extent combinations of current LLMs with text-to-image models~\citep{liu2024llm4gen, qu2023layoutllm, lian2023llm} could enable the generation of interactive worlds.

\section{Current Technical Challenges for Deep Learning in Digital Gaming: A Critical View}

While modern AI techniques, in particular deep learning, hold significant promise to enhance the future of video gaming experiences, serious technical challenges remain. As in other application areas of neural networks, these challenges frequently revolve around computational efficiency and speed, interpretability and predictability, the setting of model constraints, data requirements, model generalisation abilities, privacy considerations, and financial costs. Other, more game-specific issues centre around the complexities of integrating deep learning systems into traditional game development workflows, development time, managing player expectations with regards to AI, maintaining narrative control of video game stories, ensuring model consistency, and preserving debugging options.

One of the main concerns of game developers is the feasibility of training, running and gathering data for advanced deep learning architectures~\citep{justesen2019deep, munappy2019data}. The large amount of computation time and expensive hardware required to train state-of-the-art models such as LLMs or interactive-world generators represent an important bottleneck, especially for small and moderately-sized studios. Furthermore, if a model does not generalise effectively to new scenarios, it may have to be discarded or retrained. In supervised settings, it may be intractable to obtain sufficient amounts of human-annotated training data. If supervised data relates to in-game behaviour or player analytics, this may also potentially raise concerns regarding privacy ethics. In addition, running large trained models in real-time during gameplay could decelerate frame rate and responsiveness to unacceptably low levels. 

Further hurdles for game developers arise in connection to the \textit{black-box} nature of deep learning systems~\citep{alain2016understanding, shwartz2017opening}, which refers to the difficulty in understanding how such models make predictions and arrive at decisions. Neural networks consist of inscrutable compositions of large matrices and nonlinear functions, which makes their outputs and working mechanisms notoriously hard to interpret from a human perspective. In particular,  this makes it challenging to definitively predict the behaviour of deep networks in novel edge cases~\citep{stocco2020misbehaviour}, to guarantee their consistency, and to debug them in case they produce undesired outcomes. This opacity creates a variety of obstacles for applications in digital gaming: for instance, NPCs controlled by neural networks may exhibit unexpected and inexplicable behaviours that contradict the intended narrative or essential game mechanics; complex debugging procedures may significantly extend game development time; and it may be unclear how to hard-code necessary playability constraints into deep-learning-based procedural game level generators.

In order to reduce justified hesitancy amongst game developers to integrate novel deep learning technologies into their products, concerns like the ones outlined in this section must be addressed. Notably, a variety of currently active research areas could lead to the mitigation of some of the mentioned problems: Advances in self-supervised learning~\citep{liu2021self, chen2020big} and simulation of synthetic training data~\citep{nikolenko2021synthetic} could improve generalisation and reduce the need for labelled data sets in video game applications. More powerful model distillation~\citep{hinton2015distilling} or network pruning techniques~\citep{blalock2020state}, possibly based on the exploitation of so-called \textit{super weights} recently discovered in LLMs~\citep{yu2024super}, may reduce the computational costs associated with deep networks. Additionally, large deep learning architectures could potentially be run on distributed cloud computing systems during gameplay to offload the computational burden from local machines. Research on AI explainability~\citep{das2020opportunities} and the adversarial robustness of neural networks~\citep{silva2020opportunities} could make deep learning models more predictable, consistent, and debuggable. And quality-diversity optimisation methods~\citep{gravina2019procedural, fontaine2021differentiable} represent a growing family of algorithms that can be combined with deep learning to generate content that respects certain constraints while being diverse in nature.

\section{Conclusions}

In this work, we illuminated five promising research pathways for the application of state-of-the-art AI techniques to digital gaming: LLMs for game agent modelling, neural cellular automata for procedural content generation, deep surrogate modelling to accelerate expensive in-game simulations, self-supervised game state representation learning, and the use of unlabelled video data to train generative models of interactive worlds. The primary objective of this report is to provide a high-level overview of these areas within the current research landscape, with the aim of sparking intellectual curiosity for more targeted and in-depth research efforts in these or related fields in the future.

Video games are one of the most natural and important research frontiers in the search for general artificial intelligence systems, as they offer an almost limitless abundance of distinct cognitive tasks and simulated environments to challenge virtual agents. Further work in the areas discussed in this report could not only lead to novel technologies that enhance the quality and immersiveness of digital gaming experiences; it could also drive scientific developments in the search for more useful and capable AI models overall. For instance, research on LLM-based video game agents could lead to progress on the debated question to what extent purely language-based systems are in fact suitable as models for general intelligence~\citep{goertzel2023generative, bubeck2023sparks}; and advanced generative models of interactive worlds could supply AI agents with a potentially infinite stream of complex virtual training and testing environments~\citep{raad2024scaling, bruce2024genie}.

While we regard the research directions described in this work to hold considerable potential, it is important to note current challenges that may still limit the practical utility of deep learning in digital gaming. Issues remain around topics such as interpretability, predictability, consistency, debuggability, data requirements, generalisation, reusability, the setting of model constraints, as well as financial and computational efficiency. Systematic efforts to address these technical obstacles, alongside experimentation with novel research ideas like those outlined in this report, should be a high priority to accelerate future progress in both AI and video game research.

\newpage

\section*{Acknowledgments and Disclosure of Funding}

This exploratory report was commissioned and funded by Beam Foundation\footnote{\href{https://onbeam.com}{\faHome\ Homepage} | \href{https://x.com/BuildOnBeam}{$\mathbb{X}$}} as part of their emerging initiative to investigate the future of artificial intelligence in digital gaming.

\bibliographystyle{unsrtnat}
\bibliography{refs.bib}

\end{document}